\documentclass{article}
\usepackage{ijcai21}

\usepackage[utf8]{inputenc} 
\usepackage[T1]{fontenc}    
\usepackage{url}            
\usepackage{booktabs}       
\usepackage{amsfonts}       
\usepackage{amsmath}
\usepackage{amssymb}
\usepackage{nicefrac}       
\usepackage{microtype}      
\usepackage[pagebackref=true,breaklinks=true,colorlinks,bookmarks=false]{hyperref}

\usepackage{float}
\usepackage{times}
\usepackage{epsfig}
\usepackage{graphicx}
\usepackage{amsmath}
\usepackage{amssymb}
\usepackage[ruled,vlined]{algorithm2e}
\usepackage{multirow}
\usepackage{pgfplots}

\makeatletter
\@namedef{ver@everyshi.sty}{}
\makeatother
\usepackage{tikz}

\usetikzlibrary{shapes.geometric}
\usetikzlibrary{patterns}
\usetikzlibrary{decorations.pathreplacing}
\usetikzlibrary{calc}

\usepackage{pgfplots}
\pgfplotsset{compat=1.7}

\author{
    Konstantin Kirchheim$^1$ \and
    Tim Gonschorek$^1$ \and
    Frank Ortmeier$^1$\\
    \affiliations
    Otto-von-Guericke-University Magdeburg
    \emails
    {\{konstantin.kirchheim, tim.gonschorek, frank.ortmeier\}@ovgu.de}
}

\begin{document}

\title{Addressing Randomness in Evaluation Protocols for Out-of-Distribution Detection}
\maketitle

  \begin{abstract}
    Deep Neural Networks for classification behave unpredictably when confronted with inputs not stemming from the training distribution.
    This motivates out-of-distribution detection (OOD) mechanisms.
    The usual lack of prior information on out-of-distribution data renders the performance estimation of detection approaches on unseen data difficult.
    Several contemporary evaluation protocols are based on open set simulations, which average the performance over up to five synthetic random splits of a dataset into in- and out-of-distribution samples.
    However, the number of possible splits may be much larger, and the performance of Deep Neural Networks is known to fluctuate significantly depending on different sources of random variation.
    We empirically demonstrate that current protocols may fail to provide reliable estimates of the expected performance of OOD methods.
    By casting this evaluation as a random process, we generalize the concept of open set simulations and propose to estimate the performance of OOD methods using a Monte Carlo approach that addresses the randomness.
  \end{abstract}

  \section{Introduction}
  \label{sec:intro}
  Machine Learning driven classification systems are increasingly deployed in open, real-world environments.
      Traditionally, they operate under certain assumptions, among others, the i.i.d assumption, which asserts that samples from test and training set are mutually independent and generated by identical stationary distributions~\cite{goodfellow2016deep}, and the closed set (or closed world~\cite{boult2019learning}) assumption, which states that the data generating distribution draws from a fixed, finite set of categories~\cite{geng2020recent}.
      In more realistic scenarios, however, drifts in the data generating distribution are likely to occur between - as well as during - training and test time, and for classification tasks, the number of observable categories is (for all practical purposes) unlimited.
      Contemplating this problem,~\cite{scheirer2012toward} argues that labeling something as \emph{new, unknown or other}
      should always be considered a valid option.
      Existing literature addresses the task of recognizing inputs that do not fit into a known category (and therefore violate these assumptions) under the name of Open Set Recognition~\cite{geng2020recent}, which can be considered a special case of out-of-distribution (OOD) detection, where the distribution from which OOD samples are drawn is assumed to be conditioned on class categories~\cite{ruff2021unifying}.
      In closed-set settings where the true data generating distribution is unknown, classification models are usually selected according to the principles of empirical risk minimization~\cite{vapnik1992principles}.
      In this framework, the performance on unseen data is estimated on a subset of the available samples, which is justified by the above assumptions.
      Evaluating OOD methods turns out to be difficult because there are usually no representative OOD samples available, and the number of OOD classes is possibly infinite.
      Usually, existing datasets are adapted for this task.
      Several recent publications evaluate the performance of OOD methods by artificially splitting an existing dataset into subsets of IN and OOD classes~\cite{scheirer2012toward,scheirer2014probability,miller2021class,geng2020recent,hassen2020learning,oza2019c2ae,zhang2020hybrid}.
      The model is trained only on IN samples and tested to distinguish both IN and OOD samples - which is called \emph{open set simulation}.
      This evaluation protocol of splitting, training, and testing is repeated up to five times, and results are averaged.
      However, to our knowledge, unlike empirical risk minimization, this practice lacks a clear theoretical justification.

      Recent works demonstrated that the performance measurement of Deep Neural Networks (DNN) fluctuates significantly with several sources of random variation, among others the parameter initialization and the order in which the training data is presented to the model~\cite{bouthillier2019unreproducible}, or implementation details ~\cite{musgrave2020metric}.
      While one could try to remedy this by setting the random seed in experiments to a fixed value, this would also limit the conclusions drawn from this experiment to this particular random seed.
      \cite{bouthillier2019unreproducible} affirm that such conclusions are brittle and likely to be falsified by experiments with a different random seed.
      Instead, they recommend that deep learning experiments be replicated several times, results should be tested for statistical significance, and confidence intervals should be reported.
      The effects of randomness and the difficulty of reproducing experiments have been studied for Image Classification~\cite{bouthillier2019unreproducible}, Metric Learning~\cite{musgrave2020metric}, Image Synthesis~\cite{lucic2018gans}, and Reinforcement Learning~\cite{henderson2018deep}, yet, to our knowledge, no such study exists for OOD.

      In this work, we aim to demonstrate and address the possible issues that emerge from the inherent randomness of different OOD evaluation protocols.
      While we acknowledge the existence of protocols that draw OOD samples from unrelated datasets or synthetic distributions like Gaussian or uniform noise, this study is primarily concerned with the open set simulation framework, which is, to our knowledge, the prevalent protocol in the Open Set Recognition domain.
      In Section \ref{sec:ossim}, we provide a formal description of the concept of open set simulations that generalizes several evaluation protocols and identify several sources of randomness.
      Due to this randomness, we suspect that contemporary protocols fail to provide reliable performance estimates when comparing different OOD methods.
      In Section \ref{sec:methods},  we provide a brief overview of several OOD methods and datasets that are commonly used for comparison.
      In Section \ref{sec:randomness}, we extensively evaluate the presented methods, running three orders of magnitude more open set simulations than previous publications, and empirically demonstrate that due to significant performance fluctuations, in experiments with five-fold open set simulations, several methods have a high chance of obtaining the highest score.
      Furthermore, we provide evidence that other protocols not based on open set simulations might be subject to the same phenomenon.
      Based on this observation, we argue that the evidence provided by present evaluation protocols constitutes a brittle foundation for conclusions.
      In  Section \ref{sec:mcoss}, we propose to treat open set simulations as a fundamentally probabilistic process and cast it as a Monte Carlo approach to estimate the expected performance on unseen data.
      To the best of our knowledge, this approach is the first that systematically accounts for the performance fluctuations that stem from different sources of random variation in OOD experiments.

    \section{Open Set Simulation}
    \label{sec:ossim}
    Several existing evaluation protocols for OOD in classification tasks are based on open set simulations, meaning that they evaluate the performance of methods on a dataset by using a synthetic split of classes into in- and out-of-distribution.
    There are several variations of this protocol, some of them tailored to specific requirements of a method, which impedes comparability.
    In the following, we will formally describe the open set simulation framework, with the goal of generalizing as much as possible.
    The open set-simulations used in \cite{geng2020recent,hassen2020learning,scheirer2014probability,miller2021class,neal20180open,oza2019c2ae} can be considered a special cases of this protocol.

    OOD for classification tasks can be formulated as follows:
    Let $\mathcal{C} \subseteq \mathcal{Y}$ be a subset of all possible classes,
     $\mathcal{S} \subseteq \mathcal{X}$ a set of samples as a subset of the input domain,
     $\Psi: \mathcal{X} \rightarrow \mathcal{Y}$ a supervisor that assigns class labels, and
    $\mathcal{D} = \lbrace (x,y) \vert x \in \mathcal{S} \wedge y \in \mathcal{C} \wedge y=\Psi(x)\rbrace$ a dataset, where the $x$ are drawn from a data generating distribution.
    The goal of a method is to find a function $f: \mathcal{X} \rightarrow \mathcal{C} \cup \emptyset$ that approximates the supervisor for the codomain such that
    \begin{equation}
      f(x) =
      \begin{cases}
        \Psi(x)         \quad \text{if} \quad \Psi(x) \in \mathcal{C} \\
        \emptyset  \ \ \quad \quad \text{else} \\
      \end{cases}.
      \label{eq:ood-target-function}
    \end{equation}
    The first case of this function describes closed-set classification, while the second case introduces the OOD task.
    The open set simulation framework aims to provide a means to estimate the fidelity of a model to function (\ref{eq:ood-target-function}) without access to samples where $\Psi(x) \not \in \mathcal{C}$.

    \begin{figure}
        \centering
        \definecolor{color1bg}{HTML}{c7dfeb}
        \definecolor{color2bg}{HTML}{a1bed9}
        \definecolor{color3bg}{HTML}{6e7aad}
        \definecolor{color4bg}{HTML}{463e6e}

        \begin{tikzpicture}[
                align=center,
                scale=0.7,
                every node/.style={
                  scale=0.7
                },
                box/.style={thin,
                  minimum width=2cm, minimum height=1cm, rectangle, draw},
                kkc/.style={box,  minimum height=1cm, fill=color1bg},
                uuc/.style={box, fill=blue!20},
                unused/.style={box, pattern=north east lines},
                kuc/.style={box, fill=color3bg},
            ]
            \draw [->] (-1,1) -- node [anchor=south] {Sample Space $\mathcal{X}$} (6,1);
            \draw [->] (-1.5,0.5) -- node [anchor=south, rotate=90] {Class Space $\mathcal{Y}$} (-1.5,-4);

            \node[unused, pattern=north east lines] at (0,0.0) (train_uuc1) {};
            \node[unused, below of=train_uuc1] (train_uuc2) {};
            \node[kuc, below of=train_uuc2] (train_kuc1) {$\mathcal{D}^{out}_{train}$};
            \node[kkc, anchor=north] at (train_kuc1.south) (train_kkc) {$\mathcal{D}^{in}_{train}$};

            \node[unused, right of=train_uuc1, anchor=west] (val_uuc1) {};
            \node[kuc, below of=val_uuc1] (val_uuc2) {$\mathcal{D}^{out}_{val}$};
            \node[unused, below of=val_uuc2] (val_kuc1) {};
            \node[kkc, anchor=north] at (val_kuc1.south) (val_kkc) {$\mathcal{D}^{in}_{val}$};

            \node[kuc, right of=val_uuc1, anchor=west] (test_uuc1) {$\mathcal{D}^{out}_{test}$};
            \node[unused, below of=test_uuc1]  (test_uuc2) {};
            \node[unused, below of=test_uuc2] (test_kuc1) {};
            \node[kkc, anchor=north] at (test_kuc1.south) (test_kkc) {$\mathcal{D}^{in}_{test}$};

            \draw[decorate,decoration={brace,amplitude=5pt, raise=1mm}] let
                \p1=(test_uuc1.north east), \p2=(test_uuc1.south east)
                in ($(\x1,\y1)$) -- ($(\x2,\y2 + 1mm)$)
                node [pos=0.5, anchor=west, xshift=4mm] {$\mathcal{C}^{out}_{test}$};

            \draw[decorate,decoration={brace,amplitude=5pt, raise=1mm}] let
                \p1=(test_uuc2.north east), \p2=(test_uuc2.south east)
                in ($(\x1,\y1)$) -- ($(\x2,\y2 + 1mm)$)
                node [pos=0.5, anchor=west, xshift=4mm] {$\mathcal{C}^{out}_{val}$};

            \draw[decorate,decoration={brace,amplitude=5pt, raise=1mm}] let
                \p1=(test_kuc1.north east), \p2=(test_kuc1.south east)
                in ($(\x1,\y1)$) -- ($(\x2,\y2 + 1mm)$)
                node [pos=0.5, anchor=west, xshift=4mm] {$\mathcal{C}^{out}_{train}$};

            \draw[decorate,decoration={brace,amplitude=5pt, raise=1mm}] let
                \p1=(test_kkc.north east), \p2=(test_kkc.south east)
                in ($(\x1,\y1 - 1 mm)$) -- ($(\x2,\y2)$)
                node [pos=0.5, anchor=west, xshift=4mm] {$\mathcal{C}^{in}$};

            \draw[decorate,decoration={brace,amplitude=5pt,raise=1mm, mirror}] let
                \p1=(train_kkc.south west), \p2=(train_kkc.south east)
                in ($(\x1+1mm,\y1)$) -- ($(\x2 -1mm,\y2)$)
                  node [pos=0.5, anchor=north, yshift=-5mm] {$\mathcal{S}_{train}$};

            \draw[decorate,decoration={brace,amplitude=5pt,raise=1mm, mirror}] let
                \p1=(val_kkc.south west), \p2=(val_kkc.south east)
                in ($(\x1+1mm,\y1)$) -- ($(\x2 -1mm,\y2)$)
                  node [pos=0.5, anchor=north, yshift=-5mm] {$\mathcal{S}_{val}$};

            \draw[decorate,decoration={brace,amplitude=5pt,raise=1mm, mirror}] let
                \p1=(test_kkc.south west), \p2=(test_kkc.south east)
                in ($(\x1+1mm,\y1)$) -- ($(\x2 -1mm,\y2)$)
                  node [pos=0.5, anchor=north, yshift=-5mm] {$\mathcal{S}_{test}$};

        \end{tikzpicture}

        \caption{In open set simulations, the dataset $\mathcal{D}$ is randomly partitioned into 6 subsets.
        The model is trained on $\mathcal{D}^{in}_{train} \cup  \mathcal{D}^{out}_{train}$ and validated on $\mathcal{D}^{in}_{val} \cup \mathcal{D}^{out}_{val}$.
        Ultimately, the performance is tested on  $\mathcal{D}^{in}_{test} \cup \mathcal{D}^{out}_{test}$.
        }
          \label{img:eval-proto}
    \end{figure}
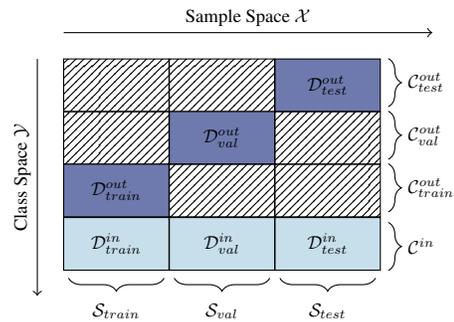

    \subsection{Dataset Construction}
    Constructing a dataset for an open set simulation comprises two substeps, which divide an existing dataset $\mathcal{D}$ into six subsets in total.
    An overview is provided in Figure \ref{img:eval-proto}. Note that some of these subsets may be empty.

    \subsubsection{Class Split}
    In the first step, the classes $\mathcal{C}$ are split into four different sets.
    One set of classes $\mathcal{C}^{in}$ whose samples are considered to be in-distribution, and three sets of out-of-distribution classes, $\mathcal{C}^{out}_{train}$, $\mathcal{C}^{out}_{val}$ and $\mathcal{C}^{out}_{test}$.
    Models will be trained on samples from $\mathcal{C}^{in}_{train}$ and potentially $\mathcal{C}^{out}_{train}$.

    \subsubsection{Sample Split}
    In a second step, the samples from $\mathcal{S}$ are divided into 3 different subsets, $\mathcal{S}_{train}$, $\mathcal{S}_{val}$ and $\mathcal{S}_{test}$ for training, validation and testing respectively.
    This aproach follows the principle of Empirical Risk Minimization for estimating the performance of supervised models on unseen data empirically.
    However, in our case, it results in a total of 6 distinct subsets of $\mathcal{D}$, such that
    $\mathcal{D}_{train}^{out} = \lbrace (x,y) \vert x \in \mathcal{S}_{train} \wedge y \in \mathcal{C}^{out}\rbrace \subseteq \mathcal{D}$, and correspondingly for the other five combinations.

    \subsection{Training}
    After dataset construction, a method is used to create a model $f$.
    This model is trained on $\mathcal{D}_{train} = \mathcal{D}^{in}_{train} \cup \mathcal{D}^{out}_{train}$.
    While there are approaches that require training or fine-tuning on out-of-distribution data (e.g.~\cite{dhamija2018reducing,hendrycks2018deep}), others do not (e.g.~\cite{hendrycks2016baseline}).
    In the latter case, $\mathcal{C}^{out}_{train}$ can be empty.
    Similarly, in some settings, validating a model before testing is not required, and $\mathcal{C}^{out}_{val}$ can be the empty set.
    However, certain techniques require a separate validation set, for example, early stopping - a method to prevent overfitting by interrupting the training once the generalization performance decreases ~\cite{prechelt1998early}. Having a separate validation set also allows tuning hyperparameters.

    \subsection{Testing}
    After the model is trained (and possibly validated), it is tested on $\mathcal{D}_{test} = \mathcal{D}^{in}_{test} \cup \mathcal{D}^{out}_{test}$.

    Different performance metrics are used to assign a performance score $\mathcal{P}(f, \mathcal{D}_{test})$ to the model.
    While the classification performance is commonly measured with the accuracy, there are several metrics for the OOD performance.
    The following two are, to our knowledge, the most frequently used:

    \paragraph{AUROC} The Area under Receiver Operating Characteristic, which characterizes the tradeoff between the false positive rate and the true positive rate, provides a threshold-independent metric for binary classification tasks.
    It ranges from zero to one, with larger values indicating better OOD performance and 0.5 corresponding to random guessing.
    \paragraph{AUPR} The Area Under Precision-Recall characterizes the tradeoff between Precision and Recall for varying thresholds. It ranges from zero to one, where larger values indicate better performance. In contrast to the AUROC, this metric is not symmetric, which means that there are two versions: one where the $\mathcal{D}^{in}_{test}$ data is treated as positive (AUPR-IN), and one where samples from $\mathcal{D}^{out}_{test}$ are treated as positive (AUPR-OUT).

    Further information regarding these metrics can be taken from~\cite{ruff2021unifying}.

    \section{Methods and Datasets}
    \label{sec:methods}
    In the following, we provide a brief overview of DNNs, OOD methods, and datasets that will be used in subsequent experiments.

    \subsection{Architecture \& Training}
    In our experiments, we employ the ResNet-18 DNN architecture \cite{he2016deep}, as it is widely used, and training is comparably cheap.
    This model comprises a 17-layered convolutional feature encoder with residual connections, followed by a global average pooling layer that produces a feature vector.
    We apply dropout  \cite{srivastava2014dropout} of 0.2 to the features and propagate the result through the final fully connected layer that outputs a logit vector $z \in \mathbb{R}^K$ where $K$ is the number of IN classes. This vector is passed through the softmax activation function
    \begin{equation}
      \sigma_i(z) = \frac{\exp(z_i)}{\sum_k^K \exp(z_k)}
    \end{equation}
    which normalizes the output for each class $i$, so that the scores can be interpreted as posterior probabilities of class membership.
    We train our models using stochastic gradient descent with a mini-batch size of 128 and weight decay of $5\times10^{-4}$ minimizing the categorical cross-entropy between the predicted class membership distribution and the one-hot encoded target label.
    We use an initial learning rate of 0.01 that we gradually vary with a cosine annealing schedule \cite{loshchilov2016sgdr}.
    We apply standard pre-processing and augmentation techniques: input normalization over $\mathcal{D}_{train}^{in}$, horizontal flipping, and random rotation and scaling.
    During training, we monitor the validation loss on $\mathcal{D}_{val}^{in}$ and interrupt the training once the loss stops decreasing.

    \subsection{Methods}

    \paragraph{Softmax Thresholding}~\cite{hendrycks2016baseline} is a baseline method for OOD. It is based on the observation that the maximum class score $\max_i \sigma_i(z)$  tends to be lower for out-of-distribution samples. A simple threshold can be applied to this value to separate in- and out-of-distribution samples.

    \paragraph{Temperature Scaling}~\cite{guo2017calibration} (TScaling) is a method that recalibrates the outputs of a DNN by dividing the logits $z$ by a constant temperature $T$ before passing them through the softmax, which leads to more uniformly distributed class probabilities. As a result, the number of over-confident predictions is reduced.

    \paragraph{ODIN}~\cite{liang2018enhancing} is a preprocessing method based on the observation that the score for the predited class $\max_i \sigma_i(z)$ changes more for in-distribution data then for out-of-distribution data when taking a single gradient step in the input space. Formally, it can be described as $\tilde{x} = x - \epsilon \ \text{sign}(-\nabla_{x} \sigma_{\hat{y}}(f_T(x)))$, where $\tilde{x}$ is the preprocessed input, $\hat{y}$ is the predicted class, $\epsilon$ is a step size, and $f_{T}$
     is the temperature scaled DNN.

    \paragraph{OpenMax}~\cite{bendale2016towards} is a post-training replacement for the softmax layer which adds an additional \emph{other} class.
    For each known class, the method estimates a cluster center in the logit space and uses a probabilistic model to calculate a pseudo-activation for the \emph{other} class. Like~\cite{oza2019c2ae}, we use the score of this class for OOD.

    \paragraph{Monte Carlo Dropout} \cite{gal2016dropout} (MCD) is a method that makes use of an interpretation of dropout as approximate Bayesian inference. While dropout is usually only applied during training of a DNN, the authors claim that when using it during inference, the average scores over multiple forward passes can be seen accurate uncertainty estimates.

    \subsection{Datasets}
    During experiments, we used both test and training set to create open set simulations from the following datasets:

    \paragraph{MNIST} 70,000 $28 \times 28$ grayscale images of 10 handwritten numbers.

    \paragraph{SVHN} 99,289 $32 \times 32$ color images of house numbering signs. The task is to determine the number on the sign; thus, there are 10 class labels in total.

    \paragraph{CIFAR-10} 60,000 $32 \times 32$ color images of 10 different animal species or objects.

    \paragraph{CIFAR-100} a larger version of the CIFAR-10 with 100 classes.
    As it still features 60,000 images, the number of per class samples is smaller than for the CIFAR-10.

    \paragraph{Tiny-ImageNet} a downscaled subset of the ImageNet dataset that with 120,000  $64 \times 64$ color images from 200 diverse classes.

  \section{Randomness in  Open Set Simulations}
  \label{sec:randomness}
  Considering the evaluation protocol described in Section \ref{sec:ossim}, we see that several steps, from the construction of an open set simulation, over the initialization of the model parameters to the training itself, are subject to randomness.
  In the case of MCD, the evaluation also involves randomness.
  As publications only recently demonstrated significant variation in the performance of DNNs based on the random seed, to the best of our knowledge, this effect has never been studied systematically in the OOD domain.
  We suspect that the effects of randomness are even more severe in this field since the nature of the evaluation introduces an additional source of random variation.

  \subsection{Impact of Class Split on Performance}
  It is usually assumed that the performance of a method varies depending on the class split since there might be configurations for which detecting OOD samples is simpler, for example, if the visual similarity between the selected known and unknown classes is large.
  Therefore, the performance is usually evaluated over several splits.
  Generally, selecting $k$ classes as $\mathcal{C}^{in}$ from a dataset with $N$ classes in total results in ${N \choose k}$ possible splits.
  If we consider the ImageNet with 1000 classes and assume 600 classes to be known, there are $ \approx 4.96 \times 10^{290}$ such splits.
  To the best of our knowledge, the common practice of using three to five splits is arbitrary and not chosen based on theoretical considerations or empirical evidence.

\begin{figure}
    \centering
    \scalebox{0.6}{\input{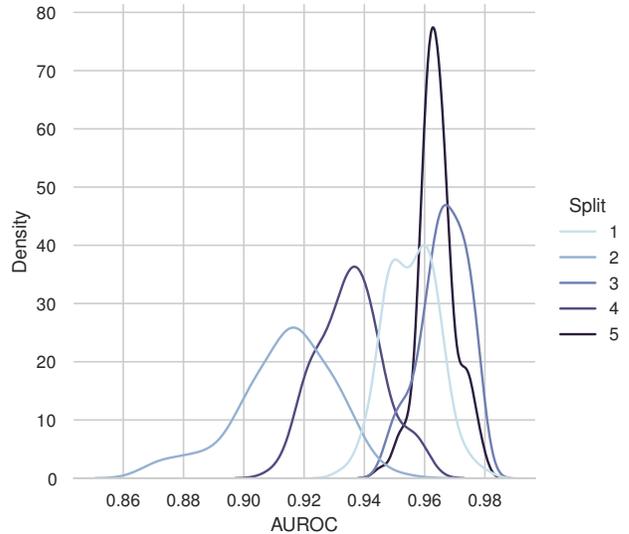}}
    \vspace{-0.7cm}
    \caption{Kernel Density Estimate of AUROC score distribution of the Softmax Thresholding baseline approach for five different class splits on the MNIST dataset.
     Experiments have been conducted over 100 different random seeds for each split.
     The performance varies significantly within, as well as between the different class splits. }
    \label{img:performance-div}
  \end{figure}

  First, we aim to determine if the class split, as usually assumed, does indeed impact the performance of methods, and if so, how this relates to other sources of randomness.
  For the five class splits of the MNIST dataset used in \cite{miller2021class}, we conducted 100 trials with different random seeds.
  The seed determined, among others, the sample split, the network parameter initialization, and the ordering of the training data.
  Figure \ref{img:performance-div} depicts the distribution of the AUROC for the Softmax Thresholding baseline approach \cite{hendrycks2016baseline}.
  As expected, the results indicate performance differences between different class splits.
  However, even for a single split, the performance varies significantly.
  Considering the observed performance variance and the vast number of possible splits, it is questionable that experiments based on five random seeds can provide reliable performance estimates.

  \subsection{Randomness in Experimental Outcomes}
  Second, we investigate if the outcomes of experiments with five open set-simulations are stable and thus able to provide a solid foundation for conclusions.
  We evaluate the OOD methods presented in Section \ref{sec:methods} on datasets with color images in 1000 open set-simulations each, which took approximately 113 hours on two Nvidia A100 GPUs.
  This number of simulations is about three orders of magnitude larger than current evaluation protocols and provides us with a large pool of possible experimental outcomes.
  For each method, we can then sample five trials from this pool and compare the average scores - which will give us a ``winning'' method.
  By iterating this process (in our case 10,000 times), we can estimate the probability that a method will win in such experiments.
  If the outcomes of such evaluations were stable, the probability of winning should be concentrated on a single method.

\begin{figure}
    \centering
    \scalebox{0.55}{\input{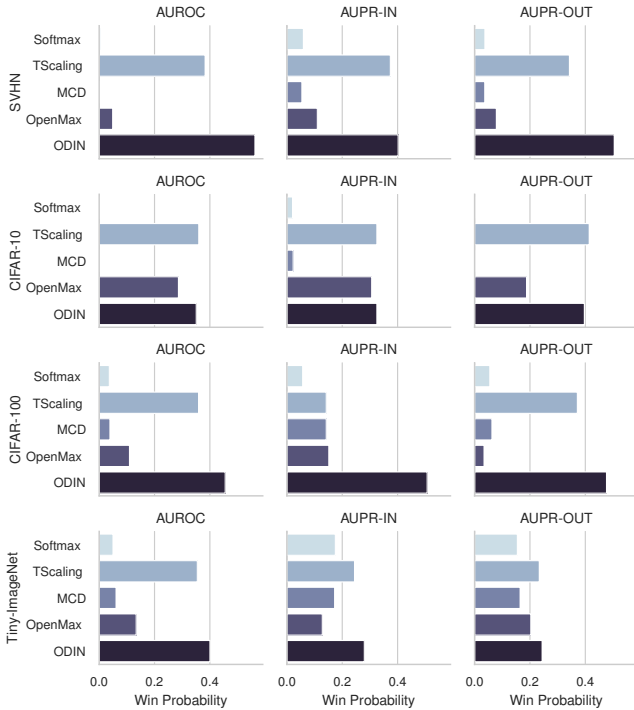}}
    \vspace{-0.8cm}
    \caption{Estimated probability of compared methods for having the highest average AUROC score over 5-fold open set simulations.
    Several methods have a high probability of winning.
    }
    \label{img:win-probs}
  \end{figure}

  The results are depicted in Figure \ref{img:win-probs}.
  The bars represent the approximate probability that a particular method will have the highest average score for the given performance metric in five open set simulations if the experiment is replicated several times with different random seeds.
  As we can see, while the newest method we tested (ODIN) usually has the highest chance of winning, the probabilities are distributed between TScaling, OpenMax, and ODIN, which means that each of these methods has a substantial chance of appearing to be the best method in such an evaluation.
  This observation holds for all datasets and across all metrics we used.
  We conclude that (with the used hyperparameters), these three methods usually outperform the baseline, as well as MCD.
  However, which of these three methods ``outperforms'' the others largely depends on random chance.
  We note that the probabilities seem to be more evenly distributed for the AUPR metrics.
  For the Tiny-ImageNet dataset and the AUPR-OUT, all methods, including the baseline, have a chance $> 15 \%$ of winning the comparison.

  \subsection{Other Evaluation Protocols}
  Apart from open set simulations, there are evaluation protocols that use samples from different datasets or synthetic data for $\mathcal{D}^{out}_{test}$~\cite{hendrycks2018deep}.
  Intuitively, this task seems easier since samples from the same dataset might share more statistical properties than samples from unrelated datasets, which would make it easier to detect the latter.

\begin{figure}
    \centering
    \scalebox{0.55}{\input{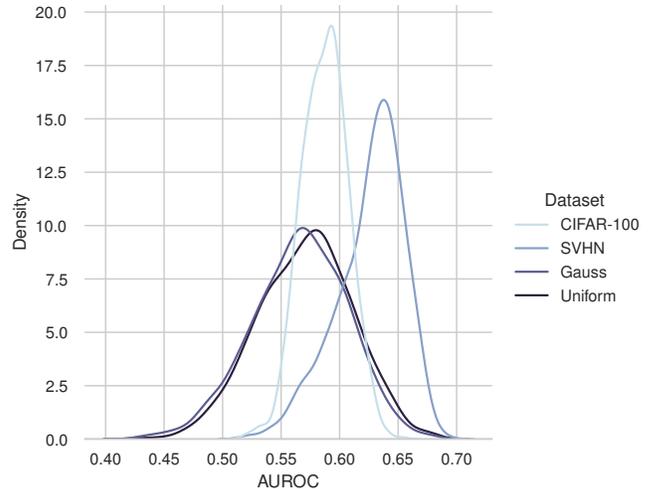}}
    \vspace{-0.7cm}
    \caption{Kernel Density Estimate of the distribution of the AUROC for the baseline approach trained on open set-simulations of CIFAR-100 and tested against $\mathcal{D}_{test}^{out}$ samples from different datasets.
    While OOD samples from the unrelated SVHN dataset yield higher AUROC scores than OOD samples from the CIFAR-100 dataset, the score of synthetic samples tends to be lower.}
    \label{img:other-eval-protos}
  \end{figure}

  Third, we conducted the following experiment: we train the baseline method on $\mathcal{D}^{in}_{train}$ from the CIFAR-100 dataset and calculated the AUROC on  $\mathcal{D}^{in}_{test}$ and samples from different datasets.
  These other datasets include the entire SVHN dataset, 1000 samples from Uniform $\mathcal{U}(0,255)$ or Gaussian Noise $\mathcal{N}(128,128)$ (clipped to $\left[0, 255 \right]$).
  Figure \ref{img:other-eval-protos} depicts the distribution of the AUROC for 1000 diffetent random seeds.
  The baseline achieves higher AUROC scores for the SVHN, which indicates that discriminating these unrelated OOD samples easier than discriminating OOD samples from the original dataset.
  For samples from Uniform and Gaussian Noise, the AUROC tends to be lower, implying that, in this setting, it can be more challenging to distinguish synthetic images than out-of-distribution samples from the same dataset.
  Further, we note that in these experiments, the performance also fluctuates significantly.
  These results show that open set simulations do not provide a lower bound for the OOD performance.
  In tests with different datasets as $\mathcal{D}^{out}_{test}$, the model might be less performant.

    \section{Probabilistic Open Set Simulation}
    \label{sec:mcoss}
    The results presented in Section \ref{sec:randomness} call the common practice of drawing conclusions based on three to five open set simulations into question.
    Just like for other fields, randomness significantly impacts the performance of OOD methods, arguably even more.
    Since all sources of random variation can be controlled in a computing environment, the execution of experiments is deterministic in theory.
    However, given the previous finding, we argue that the evaluation process should be treated as fundamentally probabilistic in practice.
    In the following, we propose a generalized version of the open set simulation framework that addresses this probabilistic behavior.
    It is conceptually simple: repeatedly sample open set simulation and evaluate methods until a sufficient confidence level is reached.

    The different sources of randomness, from the construction of an open set simulation, over the initialization of the model parameters to the training itself, can be seen as random variables and described by probability distributions.
    The score of a method $\mathcal{M}$ then becomes a function $\mathcal{O}_{\mathcal{M}}$ of this randomness.
    All randomness can be reduced to the seed that the random number generator is initialized in a deterministic computing environment.
    To compare two methods, we would compare the expected value of their score over the random seeds.

    \subsection{Monte Carlo Estimator}
    The exact calculation of the expected score is intractable because it involves a sum over all possible configurations.
    As described in Section \ref{sec:randomness}, the number of possible configurations - e.g., of class splits or parameter initializations - may be huge.
    However, the expected value can be approximated numerically, using a Monte Carlo estimate, calculated as
    \begin{equation}
      \mathbb{E}_x\left[\mathcal{O}_{\mathcal{M}}(x) \right] \approx \frac{1}{N} \sum^{N}_{i=1} \mathcal{O}_{\mathcal{M}}(x_i)
    \end{equation}
    where $N$ is the number of open set simulations, and $x_i$ is the $i$th random seed.
    This estimate will almost surely converge to the expected score for $N \rightarrow \infty$.
    For $N=5$, this approach is equivalent to the standard protocol of calculating the average score over five open set simulations.
    While this change of perspective might seem trivial, we argue that this interpretation enables the utilization of extensions of Monte Carlo methods, for example, different sampling strategies.

    \subsection{Convergence}
    In the following, we aim to provide some hints for when such a ``sufficient confidence level'' may be reached for the performance estimate.
    \cite{bouthillier2019unreproducible} proposed to run trials over a large number of random seeds and test the results for statistical significance.
    For each method and dataset, we determined the number of open set simulations required before a two-sided Welch's t-test (as used for a similar purpose by ~\cite{hassen2020learning}) with a level of significance of 0.05 would indicate a significant difference between the AUROC of the two methods. The results are depicted in Figure \ref{img:convergence}.

    We only observe six cases in which the evidence provided by five or fewer open set simulations was sufficient to infer a significant performance difference.
    For several setups, even 1000 simulations were insufficient.
    Comparing these results to Figure \ref{img:win-probs}, we note that for experiments where several methods have a high probability of winning, the number of trials required for a significant result is increased.
    Unsurprisingly, this demonstrates the effectiveness of statistical tests in preventing drawing conclusions from random outcomes.

    \begin{figure}
      \centering
      \scalebox{0.55}{\input{img/convergence.pgf}}
      \vspace{-0.7cm}
      \caption{Number of open set simulations required in our setting until a 2-sided Welch's t-test indicates a significant difference in the expected values of the AUROC for a confidence level of 0.05.
      Empty cells indicate that 1000 simulations could not provide evidence for a significant performance difference. We only find six pairings for which five or fewer iterations were sufficient.}
      \label{img:convergence}
    \end{figure}

  \subsection{Further Countermeasures}
  Our results indicate significant divergence of the OOD performance between different evaluation protocols (e.g., unknown classes from the same dataset, unrelated datasets, or synthetic samples).
  Subjecting methods to rigorous tests covering different scenarios should increase the overall robustness of conclusions against fluctuations in quantitative results.
  Since we also observe significant performance variance within different test scenarios, using a Monte Carlo approach to estimate the expected performance and testing results for statistical significance seems reasonable for other evaluation protocols as well.
  Generally, our findings illustrate limitations regarding the comparability of OOD methods tested in different settings.
  Establishing standardized test suits comprising various benchmarks could facilitate reproducibility and comparability in OOD research.

  \section{Conclusion}
In this work, we studied the effects of randomness in open set-simulation frameworks, a prevalent evaluation protocol for out-of-distribution detection.
We formalized and generalized the protocol, found that it includes several sources of randomness, and examined the effects in a large-scale study.
We ran three orders of magnitude more open set simulation than recent publications to create a pool of experimental outcomes, which we used to determine the probability that a method would appear to be the best based on the common practice of averaging the results of trials with five different random seeds.
We found that for our setting, the outcomes of open set simulations largely depend on chance.
This observation suggests that too few simulations may fail to provide a solid foundation for conclusions.
Further experiments indicate that other evaluation protocols might be subject to the same phenomenon.

Based on these observations, we proposed to treat the evaluation of OOD methods in open set simulations as a fundamentally probabilistic process and to estimate the expected value of the performance using a Monte Carlo approach to draw more reliable conclusions.
The hypothesis tests we conducted demonstrated that even a considerable number of simulations was in some cases insufficient to infer a statistically significant performance difference between the compared methods.

Future work should further investigate the performance variance of OOD evaluation protocols not based on open set simulations.
Such experiments could also include different types of data, like natural language, sound, or video.
Studying the influence of different sources of randomness in isolation might help quantify the contribution of individual factors to fluctuations and enable better design of experiments.

We want to emphasize that the goal of this work was not to evaluate the selected methods but rather to demonstrate the brittleness of a current best practice evaluation protocol.
Studies with different experimental setups or hyperparameters might come to different conclusions regarding the performance of the evaluated approaches.
However, we argue that these experiments might also be subject to inherent randomness and should address it accordingly.

  \section*{Acknowledgement}
  We thank all reviewers for their extensive and helpful feedback.

  \bibliographystyle{named}
  \bibliography{bibliography.bib}

\end{document}